\documentclass[letterpaper,10pt,conference]{ieeeconf}
\IEEEoverridecommandlockouts
\UseRawInputEncoding

\usepackage{times}
\usepackage{graphicx}
\usepackage{amsmath,amssymb,amsfonts,bm,mathtools}
\usepackage[space]{cite}
\usepackage{booktabs,multirow,array,makecell,threeparttable}
\usepackage[ruled,vlined,linesnumbered]{algorithm2e}
\usepackage{xcolor}
\usepackage{url}
\usepackage{microtype}
\usepackage[linkcolor=black,citecolor=black,urlcolor=black,colorlinks=true]{hyperref}
\usepackage{balance}
\usepackage{float}
\usepackage{url}

\newcommand{\V}{\mathcal{V}}
\newcommand{\E}{\mathcal{E}}
\newcommand{\A}{\mathcal{A}}
\newcommand{\G}{\mathcal{G}}

\newcommand{\norm}[1]{\left\lVert #1\right\rVert}
\newcommand{\relu}[1]{\left[#1\right]_{+}}
\newcommand{\method}{GuardPIBT}
\usepackage{hyperref}
\definecolor{mymagenta}{RGB}{180,0,120}

\title{\LARGE \bf
GuardPIBT: Counterfactually Gated Neural Guidance for Ultra-Large-Scale 3D Multi-Agent Path Finding
}
\author{Yuan Zhou\textsuperscript{1,6,*},
Zhenyu Hou\textsuperscript{2,*},
Guangtong Xu\textsuperscript{3},
Xiaoqiang Ji\textsuperscript{4},
Yuqing Tang\textsuperscript{5}, \\
Jialiang Hou\textsuperscript{1,6,\dag},
and Fei Gao\textsuperscript{1,6,\dag}%
\thanks{\textsuperscript{*}Indicates equal contribution.}
\thanks{\textsuperscript{\dag}Corresponding Authors:
Jialiang Hou; Fei Gao.}
\thanks{This work was supported by the National Key R\&D
Program of China under Grant No. 2023YFB4706600,
the Zhejiang Provincial Science and Technology Plan Project
under Grant No. 2024C01170,
the National Natural Science Foundation of China under
Grant Nos. 62322314 and 62203256,
and the Shenzhen Low-Altitude Airspace Strategic Program Portfolio
under Grant No. Z25306110.}
\thanks{\textsuperscript{1}Institute of Cyber-Systems and Control,
College of Control Science and Engineering,
Zhejiang University, Hangzhou 310027, China.}
\thanks{\textsuperscript{2}Huzhou Institute,
Zhejiang University, Huzhou 313000, China.}
\thanks{\textsuperscript{3}School of Automation,
Hangzhou Dianzi University, Hangzhou 310018, China.}
\thanks{\textsuperscript{4}School of Science and Engineering,
The Chinese University of Hong Kong, Shenzhen, China.}
\thanks{\textsuperscript{5}International Digital Economy Academy,
Shenzhen, Guangdong, China.}
\thanks{\textsuperscript{6}Differential Robotics Technology Company,
Hangzhou 311121, China.}
\thanks{E-mail: \{y2zhou, jlhou25, fgaoaa\}@zju.edu.cn;
xiagelearn@gmail.com; xugt@hdu.edu.cn;
jixiaoqiang@cuhk.edu.cn; ytang.cs@gmail.com.}
}

\begin{document}
\maketitle

\begin{abstract}
Large-scale 3D multi-agent path finding becomes increasingly difficult under dense traffic. Priority Inheritance with Backtracking (PIBT) scales well, but its one-step goal-directed ordering may become insufficient under dense interactions and large-scale congestion. We present GuardPIBT, which augments rather than replaces the PIBT executor: neural predictions only propose residual reorderings of PIBT's native candidates, while final actions remain determined by PIBT. First, local graph attention models nearby interactions, while global source--goal transport features provide population-level coordination context for candidate reordering. Second, a counterfactual group gate filters reorderings whose closed-loop effects may degrade coordination. Third, for ultra-large populations, population-adaptive grouping preserves decision granularity, asynchronous cached inference amortizes neural computation, and selective repair resolves long-tail agents. PIBT retains validity checking, priority inheritance, and backtracking throughout. Experiments with up to 100,000 agents demonstrate reliable completion across 2D and 3D environments, including all three 100,000-agent warehouse runs with zero audited graph violations. The project website is available at
{\color{magenta}\texttt{https://guardpibt.github.io/GuardPIBT/}}.
\end{abstract}
\section{Introduction}
\label{sec:introduction}

\begin{figure*}[t]
    \centering
    \includegraphics[width=\linewidth]{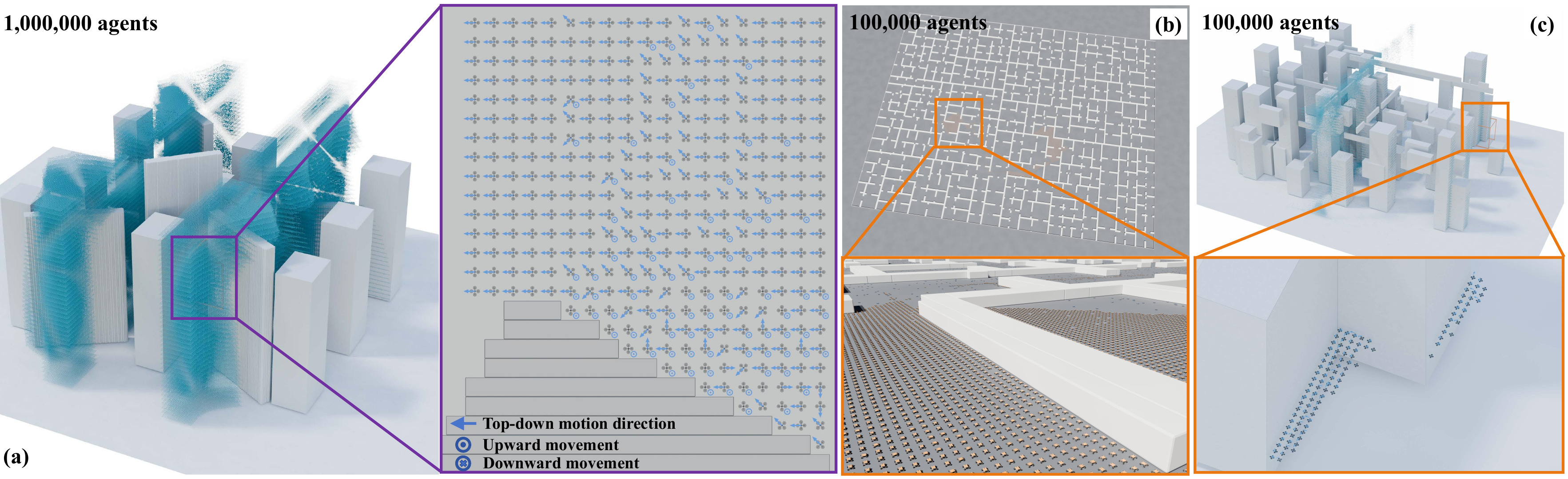}
    \vspace{-0.6cm}
    \caption{Overview of \method{} for ultra-large-scale MAPF. (a) 1000000 agents in gate obstacles. (b) 100000 agents in 2D maze. (c) 100000 agents in gate obstacles.}
    \label{fig:teaser}
    \vspace{-0.4cm}
\end{figure*}

The rapid growth of the low-altitude economy is expected to bring large fleets of delivery UAVs, inspection platforms, emergency aerial vehicles, and passenger eVTOLs into shared airspace \cite{low0}. Recent studies project future urban air mobility fleets comprising tens of thousands of vehicles, motivating coordination methods that scale beyond $10^4$ agents~\cite{asmer2025city}. These vehicles operate in complex 3D environments with dense obstacles, constrained passages, and changing origin--destination demands. Recent advances in scalable multi-robot systems further highlight the need for efficient coordination and control across heterogeneous robot teams in practical environments~\cite{ji1,ji2,ji3,mislab,argos}. Multi-agent path finding (MAPF) provides a natural low-level coordination framework by generating graph-valid joint motions without vertex or reverse-edge conflicts~\cite{ho2019utm}. As fleet sizes grow, MAPF solvers face increasing challenges in coordinating large populations while maintaining computational efficiency and reliable execution in dense environments.

Classical optimal and bounded-suboptimal MAPF solvers provide strong solution guarantees but scale poorly with the number of agents, map size, and conflict density~\cite{sharon2015cbs,barer2014ecbs}. More scalable methods improve practicality through large neighborhood search, rolling-horizon planning, and lazy successor generation over joint configurations~\cite{li2022mapflns2,li2021rhcr,okumura2023lacam}, yet still rely on repeated replanning or extensive configuration exploration. The challenge becomes more severe in 3D voxel spaces, where 26-connectivity introduces $27$ one-step candidates including waiting, and large populations create increasingly complex interaction patterns and congestion. At such scales, effective MAPF solvers require not only computational efficiency but also coordination strategies that can preserve reliable execution under dense multi-agent interactions.

PIBT and methods built on its one-step coordination mechanism provide a scalable alternative by constructing collision-free joint actions through priority inheritance and recursive backtracking~\cite{okumura2019pibt}. LaCAM scales this mechanism to $10^4$ agents by embedding PIBT-based configuration generation in an outer search, while LaGAT further introduces local graph-attention guidance~\cite{okumura2023lacam,jain2026lagat}. For lifelong MAPF, SILLM combines learned local actions, global guidance, and CS-PIBT and evaluates up to $10^4$ agents~\cite{jiang2025sillm,veerapaneni2024cspibt}. However, existing PIBT-based methods still face two important limitations in extremely dense scenarios. First, PIBT mainly relies on local interactions and goal-directed costs, which may fail to capture long-range congestion caused by many agents converging toward shared regions. Second, directly applying learned preferences is challenging because the final execution is determined by PIBT's priority inheritance and backtracking process; a preferred candidate may not necessarily improve the resulting joint motion. Therefore, effectively incorporating learning-based coordination guidance while preserving the reliability of classical MAPF executors remains a challenging problem for large-scale 3D MAPF.

We propose GuardPIBT, an executor-aligned neural enhancement framework that improves PIBT scalability without replacing its original execution mechanism. Instead of learning a standalone planner, GuardPIBT learns residual modifications to PIBT candidate ordering and allows the original PIBT executor to maintain feasibility and conflict resolution. Specifically, \textit{(i) Global--local candidate scoring:} we combine message-aware graph attention with compact global transport representations to capture both local interactions and large-scale traffic tendencies. The global representation aggregates source--goal distributions into fixed-grid transport tokens and uses rectified flow matching~\cite{liu2022rectifiedflow} to model population movement trends. Unlike map-specific global guidance graphs~\cite{chen2024trafficflow,zang2025oggo}, these tokens are generated online on a fixed 3D grid and can transfer across different maps and scales. A unified network then scores PIBT's $27$ native candidates and predicts residual ordering adjustments. \textit{(ii) Executor-aligned counterfactual gating:} Since neural preferences may not translate directly into better closed-loop execution, we introduce an action-conditioned group gate trained from paired PIBT rollouts under identical states and priority conditions. The gate evaluates whether a proposed reordering provides actual execution benefits and filters potentially harmful interventions before deployment. \textit{(iii) Ultra-large-scale execution:} To extend coordination beyond $10^4$ agents, we further introduce population-adaptive spatial grouping, asynchronous preference caching, and selective time-indexed SIPP repair to efficiently handle large-scale instances. 
Extensive experiments demonstrate that GuardPIBT achieves effective large-scale coordination, reducing runtime by up to $2.24\times$ while maintaining competitive solution costs at $10{,}000$ agents, and scaling to $1{,}000{,}000$-agent MAPF scenarios.
The main contributions are:
\begin{itemize}
    \item We propose an executor-aligned neural residual framework that enhances PIBT scalability while preserving its original collision-free execution mechanism.

    \item We introduce a global--local candidate representation that integrates local interaction modeling and global transport context for improved PIBT candidate evaluation.

    \item We develop a counterfactual group-level gating mechanism that validates neural interventions through closed-loop PIBT execution and reduces harmful candidate reorderings.

    \item We develop an ultra-large-scale MAPF extension with population-adaptive grouping, asynchronous preference caching, and selective SIPP repair, enabling robust coordination at $100{,}000$-agent and beyond scales.
\end{itemize}
\section{Related Work}
\label{sec:related}

\subsection{Scalable MAPF Solvers}

Optimal and bounded-suboptimal MAPF solvers, such as CBS and ECBS, provide strong solution guarantees but become computationally expensive as the number of agents and conflicts grows~\cite{sharon2015cbs,barer2014ecbs}. To improve scalability, later methods trade some of these guarantees for more efficient planning. MAPF-LNS2 repairs paths of selected agent subsets through large neighborhood search~\cite{li2022mapflns2}, while windowed approaches restrict multi-step planning to shorter horizons~\cite{jiang2024realistic}. LaCAM adopts lazy successor generation with PIBT-based configuration generation to achieve strong practical scalability~\cite{okumura2023lacam}.

Among lightweight MAPF solvers, PIBT performs one-step coordination through priority inheritance and backtracking~\cite{okumura2019pibt}. Its low per-step overhead makes it attractive for large agent populations, although its local candidate ordering may become insufficient under dense interactions and long-range congestion. TrafficFlow and online guidance-graph optimization address this issue by introducing global traffic-aware guidance~\cite{chen2024trafficflow,zang2025oggo}. Meanwhile, MAPF has also been extended to 3D environments and UAV traffic coordination~\cite{wang2024mapf3d,ho2019utm}, where the larger motion space further increases the difficulty of scalable coordination. These works motivate improving PIBT's candidate ordering while retaining its lightweight execution.

\subsection{Learning-Guided MAPF}

Learning-based MAPF provides another way to improve coordination by learning action preferences from local observations and inter-agent communication. PRIMAL2 learns decentralized policies through reinforcement and imitation learning, while MAGAT introduces message-aware graph attention to model neighboring interactions~\cite{damani2021primal2,li2021magat}. Since independently predicted actions may still conflict, CS-PIBT uses PIBT to convert learned preferences into collision-free one-step actions~\cite{veerapaneni2024cspibt}.
Recent work has further combined learning with scalable MAPF solvers. SILLM integrates imitation learning, global guidance, spatial communication, and CS-PIBT, and evaluates lifelong MAPF with up to $10{,}000$ agents~\cite{jiang2025sillm}. LaGAT instead introduces graph-attention guidance into LaCAM to improve configuration generation~\cite{jain2026lagat}. However, learned preferences or successor guidance may not always translate into improved closed-loop execution, since the final behavior is still determined by the underlying MAPF executor. Building on these ideas, we combine local graph-attention features with global source--goal transport context to directly score native 3D PIBT candidates, and learn candidate reordering based on execution-aware supervision rather than action preferences alone.
\section{Problem Formulation}
\label{sec:problem}

\begin{figure*}[t]
    \centering
    \includegraphics[width=0.86\textwidth]{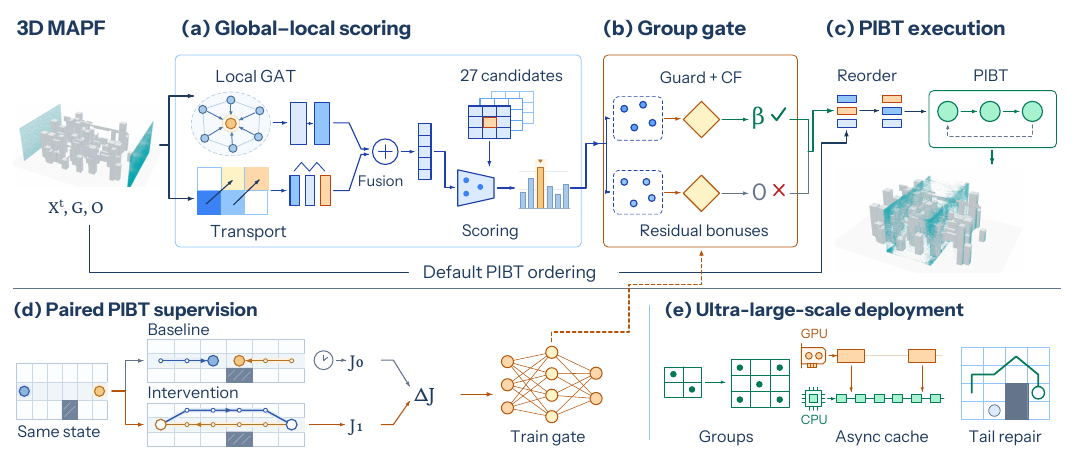}
    \vspace{-0.4cm}
    \caption{GuardPIBT pipeline: global--local candidate scoring, executor-aligned group gating, and scalable deployment with adaptive grouping, asynchronous inference, and selective tail repair.}
    \label{fig:architecture}
    \vspace{-0.6cm}
\end{figure*}

Let the free 3D voxel graph be $\G=(\V,\E)$ with dimensions $W\times H\times D$. There are $N$ labeled agents. Agent $i$ starts at $s_i\in\V$ and has goal $g_i\in\V$. At discrete time $t$, its position is $x_i^t$. The action set contains the wait action and all 26 neighboring displacements,
\begin{equation}
\A=\{-1,0,1\}^{3}, \qquad x_i^{t+1}=x_i^t+a_i^t.
\end{equation}
An action is valid if its destination is free and inside the map. We enforce the standard discrete MAPF conflicts:
\begin{align}
 x_i^{t+1} &\neq x_j^{t+1}, \label{eq:vertex_conflict}\\
 (x_i^t,x_i^{t+1}) &\neq (x_j^{t+1},x_j^t), \quad i\neq j, \label{eq:edge_conflict}
\end{align}
corresponding to vertex and reverse-edge conflicts.

We use a wait-aware Euclidean transition cost. Let $n_a=\norm{a}_0$ be the number of changed axes. Then
\begin{equation}
 c(a)=
 \begin{cases}
 1, & n_a\leq 1,\\
 \sqrt{2}, & n_a=2,\\
 \sqrt{3}, & n_a=3.
 \end{cases}
 \label{eq:motion_cost}
\end{equation}
A wait action incurs one unit of time cost, while its geometric motion length is zero. This distinction allows us to report both time-aware execution cost and pure motion distance.

The method targets online finite-horizon coordination. The horizon $T$ is used as the evaluation window for the online objective and does not indicate a fixed planning horizon; the reported experiments measure the complete execution process until termination. For an execution horizon $T$, we evaluate
\begin{equation}
J_T=\sum_{i=1}^{N}\sum_{t=0}^{T-1}c(a_i^t)
+\sum_{i=1}^{N}h_i(x_i^T),
\label{eq:completion_lb}
\end{equation}
where $h_i$ is an admissible obstacle-aware remaining-cost lower bound. We additionally report solved ratio, sum of costs for completed runs, waiting, congestion, geometric motion length, runtime, and explicit conflict counts. The online planner itself does not claim completeness or optimality.

\section{Method}
\label{sec:method}


\subsection{Overall Procedure} 
\label{sec:method_overview}
\method{} retains PIBT as the discrete executor and learns only a residual over its candidate ordering. As illustrated in Fig.~\ref{fig:architecture}, the 3D PIBT backbone first computes default candidate keys and priorities. A global--local network then combines nearby agent interactions with population-level transport context to score the same $27$ native candidates. An execution-conditioned group gate determines whether the proposed residual reordering should be activated. PIBT finally resolves claims, inheritance, displacement, and backtracking to produce the next valid joint configuration. 

Offline, the candidate scorer is trained from execution-valid PIBT states, followed by counterfactual-gate training using paired baseline/intervention PIBT rollouts. This two-stage design avoids label inconsistency caused by changing proposals during gate supervision and reduces distribution drift between learned preferences and actual PIBT execution. Online, neural preferences are updated asynchronously and cached between refreshes, while PIBT runs with the latest valid residual. For the $100{,}000$-agent setting, population-adaptive grouping and selective time-indexed repair are added without changing this core execution pipeline.

\begin{algorithm}[t]
\caption{Online Execution of GuardPIBT}
\label{alg:online}

\KwIn{Obstacles $O$, starts $S$, goals $G$, horizon $T$,
refresh interval $R$, network $f_\theta$}
\KwOut{Executed configurations $\{X^t\}_{t=0}^{T'}$}

$H \leftarrow \textsc{PrecomputeLB}(O,G)$, $X^0 \leftarrow S$\;
Initialize cached residuals $\beta_i(a)\leftarrow 0$\;

\For{$t=0,\ldots,T-1$}{
    \If{$X^t = G$}{
        \textbf{break}\;
    }

    Compute priorities $p_i^t$ and default keys $K_i^t(a)$\;

    \If{a new preference snapshot is available}{
        Build map-valid top-$k$ proposals from the latest scores\;
        
        Apply deterministic guards and group
        counterfactual gating to update $\beta_i^t(a)$\;
    }

    \If{$t \bmod R = 0$ and the provider is idle}{
        Submit $X^t$ for asynchronous neural refresh\;
    }

    Expire stale residuals and set
    $\widetilde K_i^t(a)
    \leftarrow K_i^t(a)-\beta_i^t(a)+\kappa_i^t(a)$\;

    $X^{t+1}\leftarrow
    \textsc{PIBTStep}(X^t,O,p^t,\widetilde K^t)$\;

    Update executed cost and non-goal waiting\;
}
\end{algorithm}

\subsection{Lightweight 3D PIBT Backbone}
\label{sec:method_backbone}

The learned modules require a deterministic default ordering that is fast enough to evaluate for every agent at every step. A complete 3D distance field for each labeled goal costs $O(NWHD)$ memory, so our implementation uses a relaxed but obstacle-aware lower bound and leaves one-step feasibility resolution to PIBT.

A horizontal column is blocked only when all its voxels are occupied,
\begin{equation}
O_{2\mathrm D}(x,y)=\prod_{z=0}^{D-1}O_{3\mathrm D}(x,y,z).
\label{eq:column_projection}
\end{equation}
Agents sharing the same goal column reuse one $8$-connected BFS field. Let $d_i^{xy}(x)$ be the projected distance and $d_i^z(x)=|x^z-g_i^z|$. For sorted coordinate differences $0\le d_1\le d_2\le d_3$, the obstacle-free $26$-neighbor distance and projected lower bound are
\begin{align}
h_i^{\mathrm{open}}(x)&=\sqrt3d_1+\sqrt2(d_2-d_1)+(d_3-d_2),\\
h_i^{\mathrm{proj}}(x)&=\sqrt2\min(d_i^{xy},d_i^z)+|d_i^{xy}-d_i^z|,
\end{align}
and $h_i(x)=\max\{h_i^{\mathrm{open}},h_i^{\mathrm{proj}}\}$. The implementation sorts candidates with fixed-point costs $100$, $141$, and $173$ for axial, face-diagonal, and space-diagonal moves.
For candidate $a\in\{-1,0,1\}^3$,
\begin{equation}
K_i^t(a)=c(a)+h_i(x_i^t+a).
\label{eq:default_key}
\end{equation}
The agent order uses accumulated delay and non-goal waiting,
\begin{equation}
\delta_i^t=\relu{q_i^t+h_i(x_i^t)-h_i(s_i)},\qquad
p_i^t=\delta_i^t+\lambda_ww_i^t+\epsilon_i.
\label{eq:delay_priority}
\end{equation}
PIBT itself is not modified conceptually: it processes agents by priority, tentatively claims destinations, recursively moves lower-priority occupants, releases failed claims, and backtracks. The remainder of this section describes only how candidate order is improved.

\subsection{Global--Local Candidate Scoring}
\label{sec:global_local}

Local interaction reasoning captures immediate conflicts, whereas large populations may introduce congestion patterns beyond the receptive field of neighboring agents. Meanwhile, global transport context alone cannot resolve immediate occupancy and swap risks. We therefore fuse local message-aware interactions with a source--goal-conditioned population transport representation and evaluate PIBT's exact discrete alternatives.

\subsubsection{Local interaction branch}
All coordinates are normalized by map dimensions. The implemented $13$-D agent feature is
\begin{equation}
\chi_i^t=[\bar x_i^t,\bar g_i-\bar x_i^t,
\|\bar g_i-\bar x_i^t\|_\infty,2t/T-1,v_i^{t-1},b_i^t,\deg(i)],
\end{equation}
where $v_i^{t-1}$ is the recent signed motion and $b_i^t$ indicates whether the agent is currently at its goal. Checkpoints that use accumulated 3D delay and non-goal waiting encode those two values through a separate optional delay branch. A radius graph connects at most $k_n$ nearby agents. Its directed edge feature contains the relative position normalized by the communication radius and its $L_1$ norm. Two message-aware attention layers add the edge embedding to both keys and values,
\begin{align}
s_{ji}&=\frac{q_i^\top(k_j+k_{ji}^{e})}{\sqrt{d_h}},\quad
\alpha_{ji}=\operatorname{softmax}_{j\in\mathcal N_i}s_{ji},\\
z_i^{\mathrm{loc}}&=\operatorname{FFN}\!\left(z_i+
\sum_{j\in\mathcal N_i}\alpha_{ji}(v_j+v_{ji}^{e})\right).
\label{eq:local_gat}
\end{align}
This branch represents the nearby agents most likely to compete for the same destination or participate in the same inheritance chain.

\subsubsection{Global source--goal transport branch}
Each current source--goal pair defines the rectified path
\begin{equation}
\xi_i^\tau=(1-\tau)\bar x_i^t+\tau\bar g_i,\qquad
u_i=\bar g_i-\bar x_i^t.
\label{eq:rectified_path}
\end{equation}
The implementation accepts a configurable $\tau\in[0,1]$; the reported checkpoint and online provider both use $\tau=0$. Consequently, agents are assigned to macro cells by their current positions, while the target velocity remains their normalized current-to-goal displacement. This transport branch is not used as an independent planner; instead, it provides coarse global context for candidate evaluation. The workspace is divided into a fixed $B_x\times B_y\times B_z$ grid. Every active transport or goal cell stores its center, current and goal densities, clipped density imbalance, mean remaining displacement, mean recent velocity, and $\tau$. Dense self-attention couples the active tokens, and the velocity head is supervised by
\begin{equation}
\mathcal L_{\mathrm{FM}}=\frac{1}{|\mathcal M_\tau|}
\sum_{m\in\mathcal M_\tau}\|\widehat\nu_m-\bar\nu_m\|_2^2.
\label{eq:fm_loss}
\end{equation}
Each agent queries all tokens with learned distance, relative-direction, and goal-alignment biases, producing $z_i^{\mathrm{flow}}$. The model then performs gated residual fusion,
\begin{equation}
\gamma_i=\sigma\!\left(W_g[z_i^{\mathrm{loc}};z_i^{\mathrm{flow}}]\right),\qquad
z_i=\operatorname{LN}\!\left(z_i^{\mathrm{loc}}+\gamma_i\odot z_i^{\mathrm{flow}}\right).
\label{eq:global_local_fusion}
\end{equation}
Thus, the global context is rebuilt from the current population instead of being stored as a map-specific guidance graph.

\subsubsection{PIBT-aware candidate scoring}
For every displacement, we construct
\begin{equation}
\psi_{ia}=[a_x,a_y,a_z,o_{ia},d_{ia},s_{ia},b_{ia}],
\label{eq:candidate_token}
\end{equation}
where $o_{ia}$ is current destination occupancy, $d_{ia}$ is competing default demand, $s_{ia}$ is one-step reverse-edge risk, and $b_{ia}$ identifies the default candidate. A shared network first produces an agent-conditioned $27$-action logit vector and then adds a candidate-specific score and an optional alignment with the attended transport velocity,
\begin{equation}
\ell_{ia}=[h_{\mathrm{act}}(z_i)]_a+
 w_c^\top\tanh\!\left(\phi_c(\psi_{ia})+W_qz_i\right)
 +\alpha\lambda_f\langle a,\widehat\nu_i\rangle.
\label{eq:candidate_score}
\end{equation}
The resulting $27$ logits rank the exact alternatives passed to PIBT rather than defining a separate continuous or unconstrained policy.

For candidates accepted by the counterfactual group gate, the compiled executor uses
\begin{equation}
\widetilde K_i^t(a)=10^3\!\left(K_{i,\mathrm{fix}}^t(a)-\beta_{i,\mathrm{fix}}^t(a)\right)+\kappa_i^t(a),
\label{eq:adjusted_key}
\end{equation}
where $K_{i,\mathrm{fix}}^t$ and $\beta_{i,\mathrm{fix}}^t$ are integer centi-costs and $\kappa_i^t(a)$ contains deterministic lower-order tie breaking and explicitly enabled governance terms. The learned bonus changes only the order in which PIBT tests map-valid candidates. Obstacle filtering occurs before sorting, while destination claims, parent-cell and reverse-edge rejection, priority inheritance, occupied-agent displacement, claim release, and backtracking remain inside the PIBT kernel. Setting all bonuses to zero recovers the selected PIBT baseline exactly.

\subsection{Executor-Aligned Counterfactual Group Gate}
\label{sec:counterfactual}

A high-scoring candidate does not necessarily improve the final joint
execution: PIBT may reject it or propagate its effect through a long
displacement chain. Moreover, independently accepting reorderings can produce
inconsistent interventions within the same congested region. We therefore
introduce an action-conditioned group gate that predicts whether a proposed
candidate-order intervention should be admitted before PIBT execution.

\subsubsection{Counterfactual supervision}
To obtain executor-aligned supervision, the frozen scorer generates map-valid
top-$k$ proposals from an on-policy frame, and active proposal agents are
partitioned into spatial groups. For each non-empty group $g$, baseline and
intervention rollouts start from the same configuration with identical
step-wise priority noise. The baseline preserves the default PIBT ordering,
whereas the intervention activates the group's proposed reorderings during the
first $K$ steps. Both execute the real PIBT kernel for $H$ steps, with
\begin{equation}
\begin{aligned}
J(\tau)=\;&C_{\mathrm{exec}}+H_{\mathrm{rem}}-\lambda_sN_{\mathrm{solved}}\\
&+\lambda_wW+\lambda_mL_{\mathrm{motion}}+\lambda_cC_{\mathrm{column}}.
\end{aligned}
\label{eq:cf_objective}
\end{equation}
The resulting counterfactual advantage and gate label are
\begin{equation}
A_g=J(\tau_g^{\mathrm{base}})-J(\tau_g^{\mathrm{int}}),\qquad
y_g=\mathbf1[A_g>0].
\label{eq:cf_advantage}
\end{equation}
Thus, the supervision directly reflects the closed-loop effect of the proposed
group intervention, including PIBT claims, priority inheritance, displacement,
and backtracking.

\subsubsection{Action-conditioned group gate}
The gate is a proposal-admission network rather than an action policy. It does not replace PIBT decisions; instead, it predicts whether a candidate-order modification is beneficial under closed-loop execution. For
candidate $a$ of agent $i$, it combines the fused agent representation $z_i$
with the candidate feature $\psi_{ia}$,
\begin{equation}
\ell_{ia}^{\mathrm{cf}}=\ell_i^{\mathrm{base}}+
 w_{\mathrm{cf}}^\top\tanh\!\left(W_{\mathrm{cf}}z_i+\phi_c(\psi_{ia})\right).
\label{eq:action_conditioned_gate}
\end{equation}
This design allows the same state to accept one candidate reordering while
rejecting another. During training, the logits corresponding to the proposed
candidates are aggregated within each spatial group, yielding
$\ell_g^{\mathrm{cf}}$, and optimized with
\begin{equation}
\mathcal L_{\mathrm{CF}}=
\omega(A_g)\operatorname{BCE}(\ell_g^{\mathrm{cf}},y_g)
+\lambda_\Delta|g|^{-1}\sum_{i\in g}
(\ell_{i\hat a_i}^{\mathrm{cf}}-\ell_i^{\mathrm{base}})^2,
\end{equation}
where $\omega(A_g)$ increases with $\log(1+|A_g|)$.

Online, each active proposal obtains
$p_{ia}^{\mathrm{cf}}=\sigma(\ell_{ia}^{\mathrm{cf}})$. The probabilities are
averaged within each spatial group, and its proposed reorderings are retained
only when the group mean exceeds $\tau_{\mathrm{cf}}$. Hence, the scorer
proposes candidate preferences, the counterfactual gate decides whether to
activate them, and PIBT remains responsible for final feasibility and
conflict resolution.

\subsection{Two-Stage Training}
\label{sec:training}

If the candidate scorer changes while counterfactual labels are being generated, the meaning of each intervention becomes non-stationary. Training therefore separates proposal learning from proposal admission.

In Stage~1, on-policy PIBT rollouts provide current states, default cost vectors, execution-valid target candidates, and future motion statistics. The encoder and candidate scorer minimize
\begin{equation}
\mathcal L_1=\lambda_{\mathrm{rank}}\mathcal L_{\mathrm{rank}}
+\lambda_{\mathrm{def}}\mathcal L_{\mathrm{default}}
+\lambda_{\mathrm{FM}}\mathcal L_{\mathrm{FM}}
+\lambda_{\mathrm{aux}}\mathcal L_{\mathrm{aux}}.
\end{equation}
Stage~2 freezes the transport encoder, local GAT, and candidate scorer, generates group labels using Eq.~\eqref{eq:cf_objective}, and optimizes only the counterfactual heads. The deployment threshold is selected on held-out groups using calibration and realized paired advantage. Delay-aware checkpoints additionally predict selected-action and all-candidate future waiting; these auxiliary estimates can filter likely harmful proposals but do not replace the counterfactual group gate.

\subsection{Large-Scale Deployment Extension}
\label{sec:massive_scale}

This subsection describes the additional mechanisms used only in the $100{,}000$-agent completion experiment; they are not required by the core finite-horizon policy.

\paragraph{Population-adaptive groups.}
A fixed group grid would contain $100$ times more agents when transferring from $10^3$ to $10^5$ agents. To keep the expected group population approximately constant, each axis is scaled as
\begin{equation}
B_d(N)=\max\!\left(1,\operatorname{round}\left[B_d^{\mathrm{ref}}
\left(\frac{N}{N_{\mathrm{ref}}}\right)^{1/3}\right]\right).
\label{eq:group_scaling}
\end{equation}
The reference $6\times4\times2$ grid at $N_{\mathrm{ref}}=1000$ becomes $28\times19\times9$ at $N=100{,}000$. Group probabilities use active proposal agents only, preventing completed or filtered agents from diluting the decision.

\paragraph{Sparse asynchronous refresh.}
Global transport tokens are always aggregated from the full population. The provider supports a known-active target mode in which local graph construction, candidate tensors, and output heads are built only for the selected agents; otherwise the same tensors are evaluated for all agents in fixed-size chunks. A single-flight asynchronous worker computes the next preference snapshot while PIBT reuses the previous valid bonuses. The new snapshot is installed only after completion and is cleared after a fixed time-to-live, so neural inference is amortized across many compiled PIBT steps rather than being placed on the critical path of every PIBT step.

\paragraph{Selective long-tail completion.}
Bulk coordination is handled by repeated grouped \method{} stages. When progress stalls, the implementation restarts priority noise, boosts unresolved or accepted-proposal agents, gradually relaxes the gate, and uses frontier holding to keep completed goal layers from reoccupying the remaining approach region. Only the final residual set enters time-indexed SIPP repair; settled agents become reservations and a small number of blocking goal agents may be temporarily released and returned. A short final PIBT stage reconnects the repaired paths to online execution. The final schedule is audited for vertex, reverse-edge, obstacle, and long-jump violations. This design pays the cost of time-expanded search only for the long tail, rather than for all $10^5$ agents.

\subsection{Complexity and Configuration}
\label{sec:complexity}

With $M$ active macro tokens, bounded local degree $k_n$, hidden width $d$, and neural target set $N_t$, a refresh costs
\begin{equation}
O(Nk_n d+M^2d+N_tMd+27N_td).
\end{equation}
Global token aggregation remains linear in $N$, while target-only attention and candidate scoring depend on $N_t\ll N$. Between refreshes, only the compiled PIBT kernel runs. Shared goal columns reduce lower-bound storage from $O(NWHD)$ to $O(|\mathcal U|WH)$.

The reference network uses hidden width $64$, four heads, two local GAT layers, one token-attention layer, a $12\times8\times4$ macro grid, and at most $16$ graph neighbors. Counterfactual labels use $H=32$ rollout steps and a $K=8$-step intervention. The standard scorer returns top-$k=3$ proposals, while the frozen $100{,}000$-agent profile may use top-$k=1$ together with a centi-cost bonus. The optional completion extension obtains empirical full completion through selective repair rather than full-population configuration search.


\section{Experimental Evaluation}
\label{sec:experiments}


\subsection{Experimental Setup}
\label{sec:exp_setup}
We evaluate \method{} in terms of scalability, cross-scene generalization, and component effectiveness. Experiments are conducted on an AMD EPYC 7502 CPU, 1~TiB of system memory, and eight NVIDIA RTX~4090 GPUs. Our solver does not rely on multi-GPU parallelism: the large-scale audited runs use a single RTX~4090 together with CPU parallelism. The PIBT executor is implemented in C++17, while neural modules use Python, PyTorch, and CUDA.

We compare \method{} with LaCAM~\cite{okumura2023lacam},
LaGAT~\cite{jain2026lagat}, and PyPIBT~\cite{okumura2019pibt}
from $100$ to $100{,}000$ agents. We then evaluate
$100{,}000$-agent generalization across different scenes and perform
matched Warehouse-3D ablations.
We report end-to-end time $T_{\mathrm{E2E}}$, average sum of costs $\bar{C}_{\mathrm{SOC}}=C_{\mathrm{SOC}}/N$, where $C_{\mathrm{SOC}}$ is the sum of all agent path costs, and the average number of waiting steps per agent $\bar{W}_{\mathrm{step}}$.

\subsection{Scalability Across Agent Populations} 
\label{sec:scaling} 
To evaluate how GuardPIBT scales with increasing agent populations, we compare its $T_{\mathrm{E2E}}$ and $\bar{C}_{\mathrm{SOC}}$ against representative baselines from 100 to 10,000 agents. The goal of this experiment is to examine whether our method can maintain a favorable $T_{\mathrm{E2E}}$--$\bar{C}_{\mathrm{SOC}}$ trade-off as coordination becomes increasingly difficult. Table~\ref{tab:scaling} summarizes the results on Forest, Maze, and Warehouse scenes. 

At small scales, GuardPIBT introduces additional inference and coordination overhead and is therefore slower than lightweight search-based methods such as LaCAM. Nevertheless, it achieves the lowest $\bar{C}_{\mathrm{SOC}}$ across all three scenes at $N=100$, showing that the proposed method can obtain high-quality solutions even when $T_{\mathrm{E2E}}$ scalability is not yet the dominant challenge. As the population increases, the overall advantage becomes more evident. At $N=10{,}000$, GuardPIBT provides complete reported $T_{\mathrm{E2E}}$ and $\bar{C}_{\mathrm{SOC}}$ results across all three scenes. In Forest, it requires $172.41$~s with a $\bar{C}_{\mathrm{SOC}}$ of $241.33$, compared with $386.40$~s and $249.77$ for LaGAT, yielding a $2.24\times$ speedup with slightly lower $\bar{C}_{\mathrm{SOC}}$. In Maze, GuardPIBT reduces $T_{\mathrm{E2E}}$ from $174.89$~s to $113.53$~s relative to LaCAM while reducing $\bar{C}_{\mathrm{SOC}}$ from $2910.90$ to $2051.33$, corresponding to a $1.54\times$ speedup and a $29.5\%$ $\bar{C}_{\mathrm{SOC}}$ reduction. 

GuardPIBT does not dominate every metric. In the $10{,}000$-agent Warehouse scene, LaCAM is faster, while LaGAT achieves a lower $\bar{C}_{\mathrm{SOC}}$. However, GuardPIBT remains substantially faster than PyPIBT and LaGAT while maintaining competitive solution quality. These results show that the main strength of GuardPIBT lies in balancing computational efficiency and solution quality as the agent population grows, rather than optimizing a single metric in every setting.

\begin{table*}[t]
\centering
\caption{Scalability comparison on 2D Forest, Maze, and Warehouse scenes.}
\label{tab:scaling}
\footnotesize
\renewcommand{\arraystretch}{0.95}
\newcommand{\metrics}{\makecell[c]{$T_{\mathrm{E2E}}(\mathrm{s})$\\[-0.15ex]$\bar{C}_{\mathrm{SOC}}$}}
\newcommand{\metricpair}[2]{\makecell[c]{#1\\[-0.15ex]#2}}
\newcommand{\best}[1]{\textbf{#1}}

\resizebox{\textwidth}{!}{%
\begin{tabular}{@{}ll|cccc|cccc|cccc@{}}
\toprule
\multirow{2}{*}{Scene} & \multirow{2}{*}{Metric}
& \multicolumn{4}{c|}{$N=100$}
& \multicolumn{4}{c|}{$N=1{,}000$}
& \multicolumn{4}{c}{$N=10{,}000$} \\
\cmidrule(lr){3-6}\cmidrule(lr){7-10}\cmidrule(lr){11-14}
& & Guard PIBT & LaCAM & PyPIBT & LaGAT
  & Guard PIBT & LaCAM & PyPIBT & LaGAT
  & Guard PIBT & LaCAM & PyPIBT & LaGAT \\
\midrule
Forest & \metrics
& \metricpair{6.24}{\best{18.93}}
& \metricpair{\best{0.02}}{24.58}
& \metricpair{0.28}{24.36}
& \metricpair{8.12}{20.78}
& \metricpair{14.31}{\best{63.68}}
& \metricpair{\best{0.49}}{91.96}
& \metricpair{--}{--}
& \metricpair{10.75}{72.74}
& \metricpair{\best{172.41}}{\best{241.33}}
& \metricpair{--}{--}
& \metricpair{--}{--}
& \metricpair{386.40}{249.77} \\
\midrule
Maze & \metrics
& \metricpair{6.24}{\best{26.82}}
& \metricpair{\best{0.02}}{36.98}
& \metricpair{0.28}{35.44}
& \metricpair{8.05}{28.39}
& \metricpair{16.36}{275.61}
& \metricpair{\best{0.57}}{289.90}
& \metricpair{178.17}{374.69}
& \metricpair{54.21}{\best{224.24}}
& \metricpair{113.53}{\best{2051.33}}
& \metricpair{\best{174.89}}{2910.90}
& \metricpair{--}{--}
& \metricpair{--}{--} \\
\midrule
Warehouse & \metrics
& \metricpair{6.61}{\best{22.06}} & \metricpair{\best{0.02}}{28.47}
& \metricpair{0.29}{29.18} & \metricpair{8.04}{23.75}
& \metricpair{8.96}{\best{70.72}} & \metricpair{\best{0.38}}{99.58}
& \metricpair{12.378}{99.67} & \metricpair{10.269}{77.38}
& \metricpair{168.936}{315.67} & \metricpair{\best{39.64}}{394.87}
& \metricpair{888.188}{394.93} & \metricpair{257.24}{\best{288.04}} \\
\bottomrule
\end{tabular}%
}
\vspace{0.5mm}
\begin{minipage}{\textwidth}
\scriptsize
Each cell lists mean $T_{\mathrm{E2E}}$ and mean $\bar{C}_{\mathrm{SOC}}$ from top to bottom.
$T_{\mathrm{E2E}}$ is averaged over both attempts; $\bar{C}_{\mathrm{SOC}}$ is averaged over complete, audited solutions only.
Bold indicates the lowest available value within the same scene and population size.
A dash indicates unavailable complete-solution $\bar{C}_{\mathrm{SOC}}$.
\vspace{-0.4cm}
\end{minipage}
\end{table*}

\subsection{Large-Scale Cross-Scene Generalization}
\label{sec:cross_scene}

To evaluate whether GuardPIBT can handle substantially larger MAPF instances while retaining robustness across different environments, we further test the method on diverse 2D and 3D scenes with $100{,}000$ agents.
This experiment is designed to examine both its extreme-scale coordination capability and its generalization across different map structures, obstacle distributions, and dimensionalities.
Table~\ref{tab:100k_cross_scene} reports results on Forest-2D, Maze-2D, Warehouse-2D, Gate-Obstacle-2D, Gate-Walls-3D, and Warehouse-3D, without scene-specific adaptation.
Representative cross-scene results are shown in Figs.~1,~3 and ~4, while Fig.~5 illustrates the execution process of a $100{,}000$-agent Warehouse-3D run from initialization to final completion.

The $T_{\mathrm{E2E}}$ varies substantially with scene structure.
Gate-Walls-3D is solved in $194.61$~s, while the more constrained Maze-2D requires $5208.15$~s.
Warehouse-3D is completed in $908.61$~s, and the remaining 2D scenes require approximately $1105$--$1475$~s.
Despite these differences, GuardPIBT completes all reported settings, demonstrating that its large-scale coordination capability remains effective across substantially different environments rather than being specialized to a particular map topology.
The snapshots in Fig.~4 further show how the $100{,}000$-agent Warehouse-3D instance evolves through dense intermediate coordination before reaching completion.

We further test a one-million-agent 3D instance, shown in Fig.~1(a).
All $1{,}000{,}000$ agents reach their goals without audited vertex, swap, or obstacle collisions, with an end-to-end $T_{\mathrm{E2E}}$ of approximately $995$~s.
This result further demonstrates that GuardPIBT can scale beyond the $100{,}000$-agent regime while preserving valid coordination in extremely large MAPF instances.

\begin{figure}[t]
 \centering
    \includegraphics[width=0.9\linewidth]{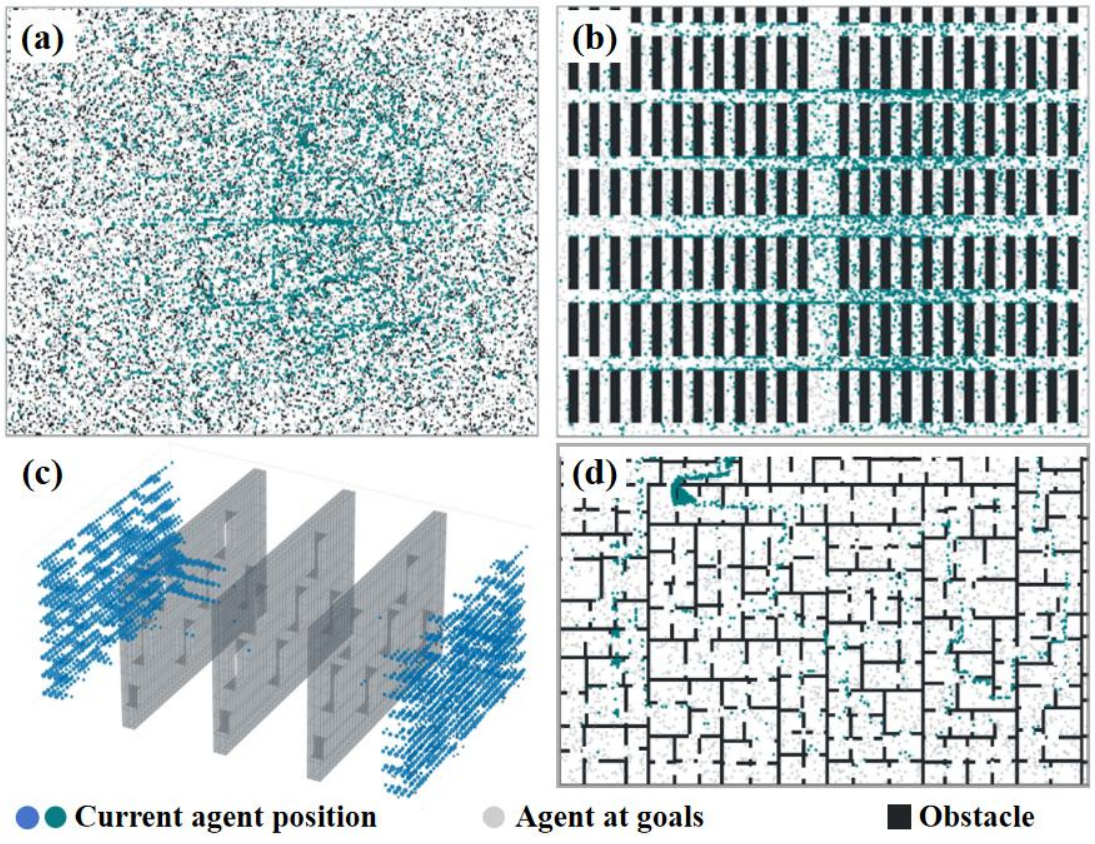}
    \vspace{-0.4cm}
\caption{Cross-scene generalization of \method{} across diverse 2D and 3D environments. (a)–(d) correspond to a 2D random forest, 2D warehouse, 3D gate walls, and maze, respectively.}
\label{fig:scene_generalization}
\end{figure}

\begin{figure}[t]
 \centering
    \includegraphics[width=0.9\linewidth]{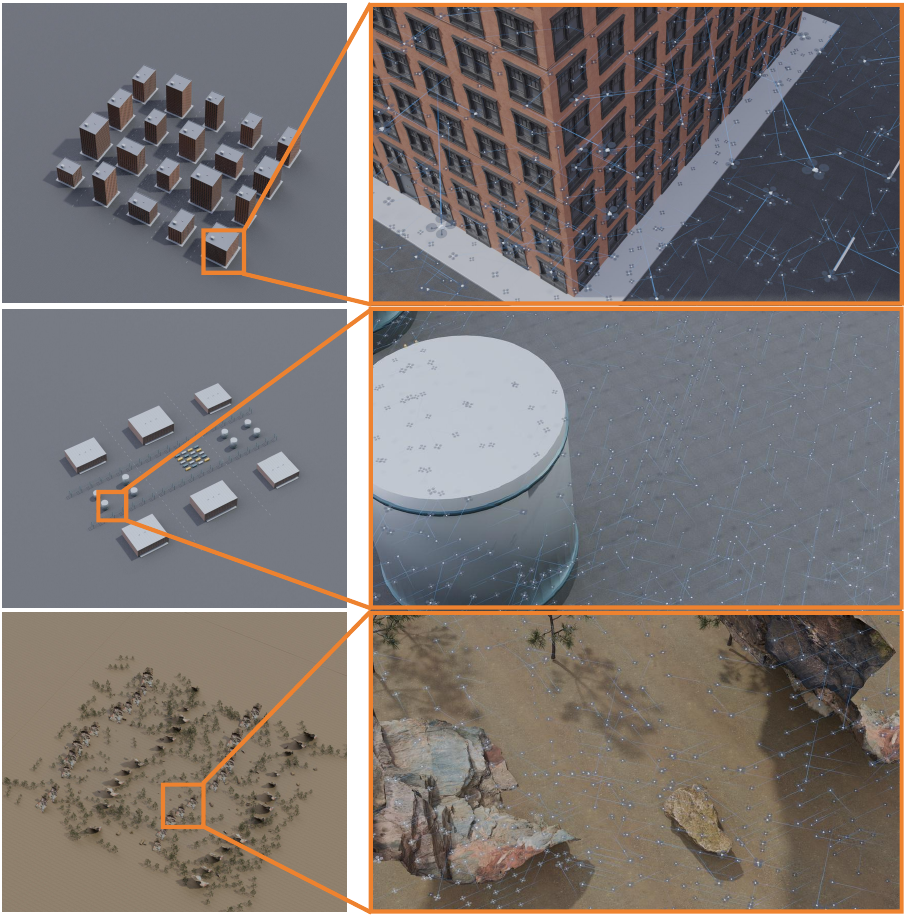}
    \vspace{-0.4cm}
\caption{Cross-scene generalization of \method{}}
\label{fig:scene_generalization2}
\end{figure}

\begin{figure*}[t]
 \centering
    \includegraphics[width=\linewidth]{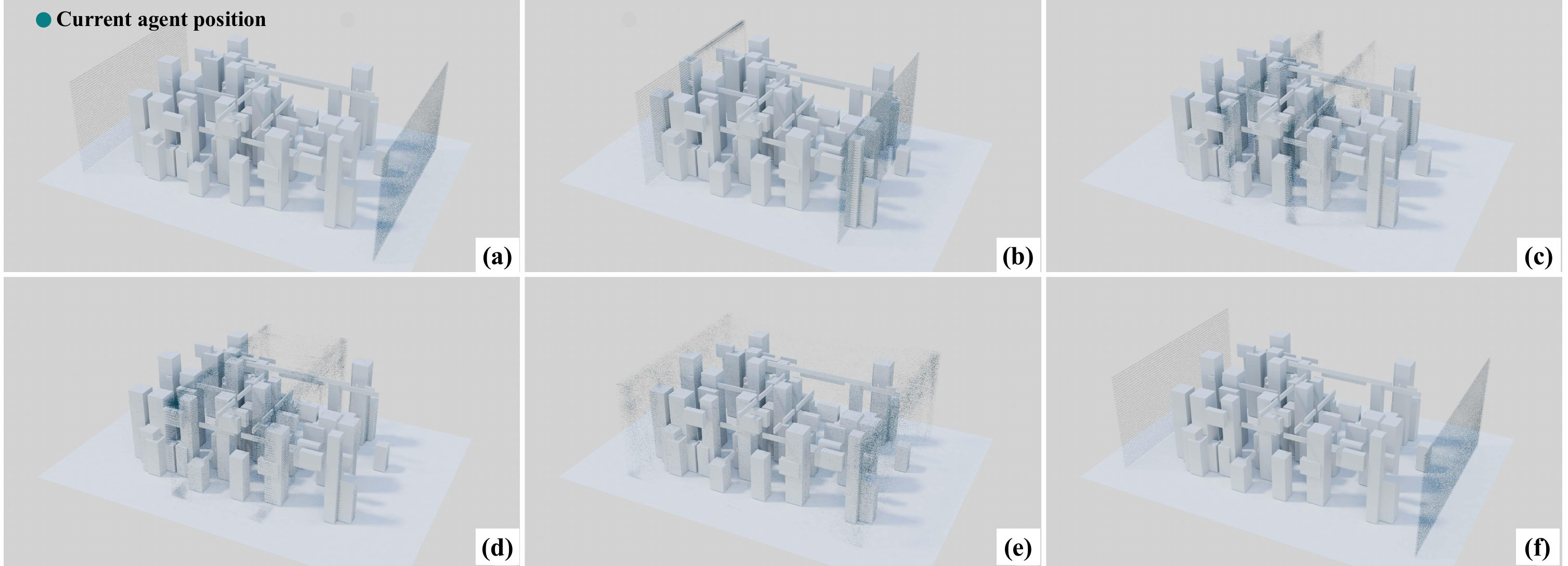}
    \vspace{-0.4cm}
\caption{Snapshots of a $100{,}000$-agent Warehouse-3D run from initialization through dense coordination to final completion.}
\vspace{-0.4cm}
\label{fig:scene_generalization_3d}
\end{figure*}

\begin{table}[t]
\centering
\caption{100k-agent cross-scene results. F/M/W denote Forest/Maze/Warehouse; GW denote Gate Walls.}
\vspace{-0.2cm}
\label{tab:100k_cross_scene}
\scriptsize
\setlength{\tabcolsep}{3.2pt}
\renewcommand{\arraystretch}{1.05}
\resizebox{\columnwidth}{!}{
\begin{tabular}{lcccccc}
\toprule
Metric
& F-2D
& M-2D
& W-2D
& GW-3D
& W-3D \\
\midrule
$T_{\mathrm{E2E}}$ (s)
& 1104.83 & 5208.15 & 1163.11& 194.61  & 908.61 \\
$\bar{C}_{\mathrm{SOC}}$
& 2573.77 & 16558.92 & 2228.62 & 659.31  & 1041.83 \\
\bottomrule
\end{tabular}
}
\end{table}


\subsection{Core Component Ablation}
\label{sec:ablation}

To examine whether the learned components improve closed-loop coordination rather than merely introducing additional inference overhead, we compare the full model with variants that remove the global transport branch or the counterfactual group gate.
Table~\ref{tab:core_ablation} reports the results on the $10{,}000$-agent 3D Warehouse setting.
All variants complete all 10 evaluated runs, allowing their coordination quality to be compared without conflating lower $\bar{C}_{\mathrm{SOC}}$ with failed.

Removing the global flow branch increases the average SOC from $3739.17$ to $4089.50$ and the average waiting steps per agent from $2192.83$ to $2521.93$, while the waiting ratio rises from $58.52\%$ to $61.54\%$.
This indicates that population-level transport context complements local interaction reasoning by reducing inefficient motion and prolonged waiting.
Removing the counterfactual gate produces a smaller increase in SOC and average waiting steps, from $3739.17$ to $3772.39$ and from $2192.83$ to $2205.99$, respectively.
Its $T_{\mathrm{E2E}}$ and waiting ratio remain close to those of the full model, suggesting that the gate mainly improves the quality of learned candidate interventions rather than reducing computation.

Overall, the relatively small $T_{\mathrm{E2E}}$ differences among the three variants show that neither component is designed primarily for $T_{\mathrm{E2E}}$ acceleration.
Instead, the global transport branch provides population-level coordination context, while the counterfactual gate filters candidate reorderings according to their predicted closed-loop effect under PIBT execution.

\begin{table}[t]
\centering
\caption{Core mechanism ablation on the $10{,}000$-agent 3D warehouse.}
\vspace{-0.2cm}
\label{tab:core_ablation}
\footnotesize
\renewcommand{\arraystretch}{1.05}

\begin{tabular*}{\columnwidth}{@{\extracolsep{\fill}}lccc@{}}
\toprule
Variant
& $T_{\mathrm{E2E}}$ (s)
& $\bar{C}_{\mathrm{SOC}}$
& $\bar{W}_{\mathrm{step}}$ \\
\midrule

\method{}
& 256.10
& \textbf{3739.17}
& \textbf{2192.83} \\

W/o global flow
& 251.45
& 4089.50
& 2521.93 \\

W/o CF gate
& \textbf{250.34}
& 3772.39
& 2205.99 \\

\bottomrule
\end{tabular*}
\vspace{1cm}
\begin{minipage}{\columnwidth}
\scriptsize
All variants complete all 10 evaluated runs.
The no-global-flow variant uses the legacy deployment setting.
\vspace{-1cm}
\end{minipage}
\end{table}
\section{Conclusion}
\label{sec:conclusion}
We presented GuardPIBT, an executor-aligned neural enhancement of PIBT for ultra-large-scale 3D MAPF. By combining local graph attention with population-level transport context, GuardPIBT learns residual candidate reorderings, while a counterfactual group gate filters modifications that are unlikely to improve closed-loop execution. Population-adaptive grouping, asynchronous cached inference, and selective tail repair further extend the framework to extreme populations. Experiments across diverse 2D and 3D scenes demonstrate favorable runtime--cost trade-offs, reliable completion at $100{,}000$ agents, and successful million-agent execution with zero audited graph violations. Future work will extend GuardPIBT toward continuous-time and kinodynamic coordination for large-scale aerial robot systems.


\bibliographystyle{./support/IEEEtran}
\bibliography{references}
\end{document}